\documentclass[10pt, conference]{IEEEtran}
\IEEEoverridecommandlockouts
\usepackage{cite}
\usepackage{amsmath,amssymb,amsfonts}
\usepackage{algorithmic}
\usepackage{graphicx}
\usepackage{textcomp}
\usepackage{xcolor}
\usepackage[utf8]{inputenc}
\usepackage{CJK}
\usepackage{comment}
\usepackage{subfigure}
\usepackage{float}
\usepackage{algorithm}  
\usepackage{amsmath}  
\usepackage{mathrsfs}
\usepackage{amsmath}
\usepackage{amssymb}
\usepackage{makecell}
\usepackage{multirow}
\newcommand{\tabincell}[2]{\begin{tabular}{@{}#1@{}}#2\end{tabular}}
\usepackage{soul}
\usepackage{textcomp} 
\usepackage{bm}
\usepackage{booktabs}
\usepackage{mathrsfs}
\usepackage{amsmath}
\usepackage{marvosym}
\usepackage{etoolbox}

\makeatletter
\patchcmd{\@makecaption}
  {\scshape}
  {}
  {}
  {}
\makeatletter
\patchcmd{\@makecaption}
  {\\}
  {.\ }
  {}
  {}
\makeatother

\begin{document}
\title{
BERT2DNN: BERT Distillation with Massive Unlabeled Data for Online E-Commerce Search
}

% put a textsection to footnote
% https://tex.stackexchange.com/questions/520691/footnote-with-same-symbol-for-different-authors

\author{
\IEEEauthorblockN{
Yunjiang Jiang\textsuperscript{\textsection}\IEEEauthorrefmark{1},
Yue Shang\textsuperscript{\textsection}\IEEEauthorrefmark{1},
Ziyang Liu\textsuperscript{\textsection}\IEEEauthorrefmark{1}\textsuperscript{\Letter}, 
Hongwei Shen\IEEEauthorrefmark{1}, 
Yun Xiao\IEEEauthorrefmark{1}, \\
Wei Xiong\IEEEauthorrefmark{1},  
Sulong Xu\IEEEauthorrefmark{1}, 
Weipeng Yan\IEEEauthorrefmark{1},
Di Jin\IEEEauthorrefmark{2}
}
\IEEEauthorblockA{\IEEEauthorrefmark{1}JD.com, China}
\IEEEauthorblockA{\IEEEauthorrefmark{2}College of Intelligence and Computing, Tianjin University, China}
\IEEEauthorblockA{\{yunjiang.jiang,yue.shang,liuziyang7\}@jd.com}
}
\maketitle
\begingroup\renewcommand\thefootnote{\textsection}
\footnotetext{These authors contributed equally to this research.}
\endgroup

\begin{abstract}
Relevance has significant impact on user experience and business profit for e-commerce search platform. In this work, we propose a data-driven framework for search relevance prediction, by distilling knowledge from BERT and related multi-layer Transformer teacher models into simple feed-forward networks with large amount of unlabeled data. The distillation process produces a student model that recovers more than 97\% test accuracy of teacher models on new queries, at a serving cost that's several magnitude lower (latency 150x lower than BERT-Base and 15x lower than the most efficient BERT variant, TinyBERT). The applications of temperature rescaling and teacher model stacking further boost model accuracy, without increasing the student model complexity.

We present experimental results on both in-house e-commerce search relevance data as well as a public data set on sentiment analysis from the GLUE benchmark. The latter takes advantage of another related public data set of much larger scale, while disregarding its potentially noisy labels. Embedding analysis and case study on the in-house data further highlight the strength of the resulting model. By making the data processing and model training source code public, we hope the techniques presented here can help reduce energy consumption of 
the state of the art Transformer models and also level the playing field for small organizations lacking access to cutting edge machine learning hardwares.
\end{abstract}


\section{Introduction}
\begin{CJK}{UTF8}{gbsn}
Search engine is of critical importance to e-commerce business since it's the most direct way for users to access items of interest. Within search ranking, relevance of retrieved results has always been a core challenge. Users may easily quit or rephrase their queries when they see a significant number of irrelevant results. While the scope of e-commerce search is significantly less than that of web-search, there are additional considerations specific to the business needs, such as key business metrics like GMV (general merchandise value), CTR (click through rate), and CVR (conversion rate). Moreover, e-commerce search engine offers other result sorting options, such as Price Low to High, best sellers, which may place non-relevant items high in the results list. Thus \emph{relevance prediction} models play a key role in inducing user clicks and conversions (purchase after click) for e-commerce search and face additional challenges in serving costs.

\begin{figure}[!ht]
\centering
\subfigure[Traditional model]{
\includegraphics[width=0.16\textwidth]{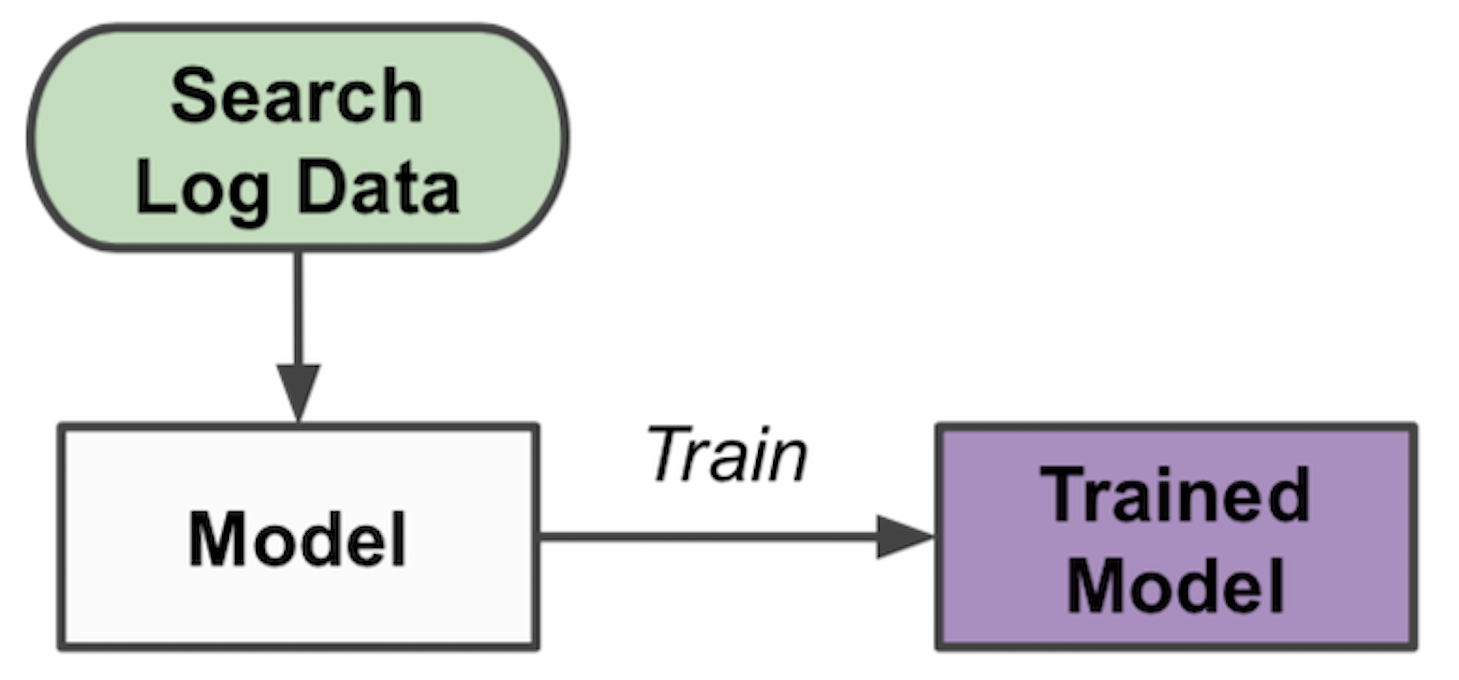}
\label{Relation to previous work1}}
\hspace{0in}
\subfigure[Similar work \cite{jiang2019unified}]{
\includegraphics[width=0.26\textwidth]{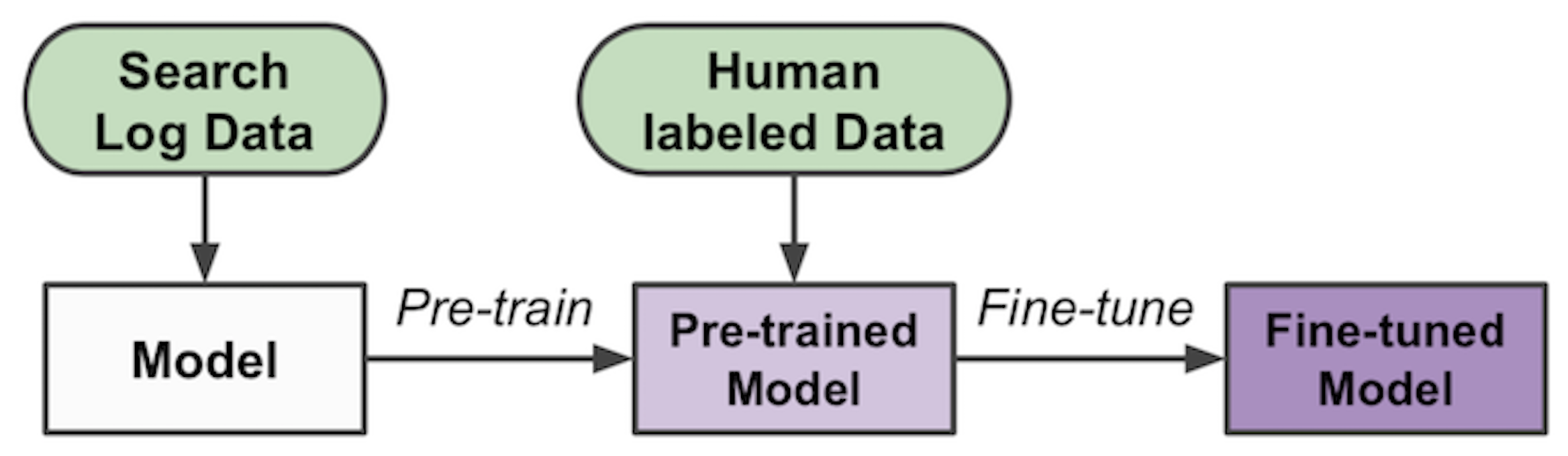}
\label{Relation to previous work2}}
\hspace{0in}
\subfigure[Our method: BERT2DNN]{
\includegraphics[width=0.42\textwidth]{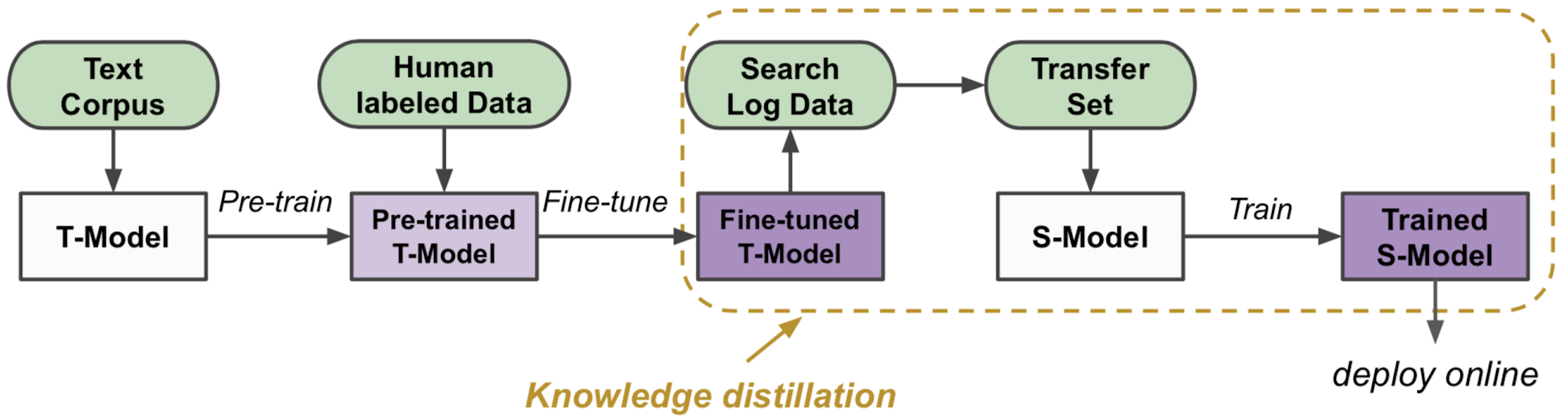}
\label{Relation to previous work3}}
\hspace{0in}
\caption{Relation to previous work. T-model represents teacher model, S-model represents student model. The content in the dashed rectangle is knowledge distillation which will be expounded in Section 3.}
\end{figure}

In the online environment, conventional e-commerce search engine mainly includes the following three phases: retrieval, ranking, and re-ranking. The task of relevance prediction is usually placed in the phase of ranking, which rearranges results based on their relevance to the query. In offline experiments, relevance models are often trained using accurate human-labeled data. Unfortunately, due to high labeling costs, only a small amount of human-labeled data is available, severely limiting the model's ability to learn. Some traditional CTR or CVR prediction methods (Figure \ref{Relation to previous work1}) treat user behavior such as click or purchase as proxies of true relevance labels, but this approach quickly reaches its limitation in accuracy. Because the click or purchase signal is noisy and affected by many kinds of behavioral biases (e.g. position bias\cite{joachims2017accurately} and presentation bias\cite{mao2018constructing}), it is systematically different from true relevance.

In the present work we attempt to overcome the above difficulties by applying the techniques of transfer learning and knowledge distillation to improve result relevance in the e-commerce setting. Recently, neural transfer learning has shown its strong ability in computer vision, natural language processing and other fields \cite{devlin2019bert, gamrian2018transfer, zhang2019ernie}. Its core idea is to extract knowledge from source tasks and apply it to improve target task's performance. It is worth noting that even though transfer learning (Figure \ref{Relation to previous work2}) has been applied in the context of e-commerce search \cite{jiang2019unified}, two outstanding difficulties remain: 1) during pre-training, the model such as \cite{jiang2019unified} relies on user behavior as label information, which often contradicts the relevance objective; 2) during fine-tuning, the scarcity of editorially labeled relevance data (in the order of hundreds of thousands) makes it difficult for the model to capture relevance in all e-commerce search scenarios. Another important technique is knowledge distillation, whereby a highly effective teacher model produces predictions on a dataset as labels for a student model of typically simpler architecture to learn from. The invention of BERT, and its subsequent improvements, precipitated amazing progress in both areas above. The pre-trained BERT model is suitable for myriad downstream tasks, even with tiny amounts of data labeling. With the dramatic improvement in model accuracy, however, the model complexity and serving costs also skyrocketed. While big companies like Google have developed highly efficient in-house hardware to serve the expensive BERT model directly \cite{jouppi2017datacenter, nayak_2019}, most companies struggle to bring it to production despite its ease of training and quality gain. 

Here we present \textbf{BERT2DNN}, a practical solution to distill BERT down to a bare-bone feed-forward neural network of 4 layers, without sacrificing much of the model quality. The main steps are visually summarized in Fig. \ref{Relation to previous work3}, where T-model stands for the BERT based single or ensemble teacher model, and S-model the simple feed-forward student model (the key modeling innovations, enclosed in the dashed rectangle, will be expounded in Section 3). The student model uses prediction scores of fine-tuned BERT model as label, whose relevance accuracy is high. The transfer set is large and diverse enough to cover multiple queries and items. As a result, BERT2DNN successfully addresses the problems in Fig. \ref{Relation to previous work2}.

\begin{figure}[!ht]
\centering
\includegraphics[width=0.48\textwidth]{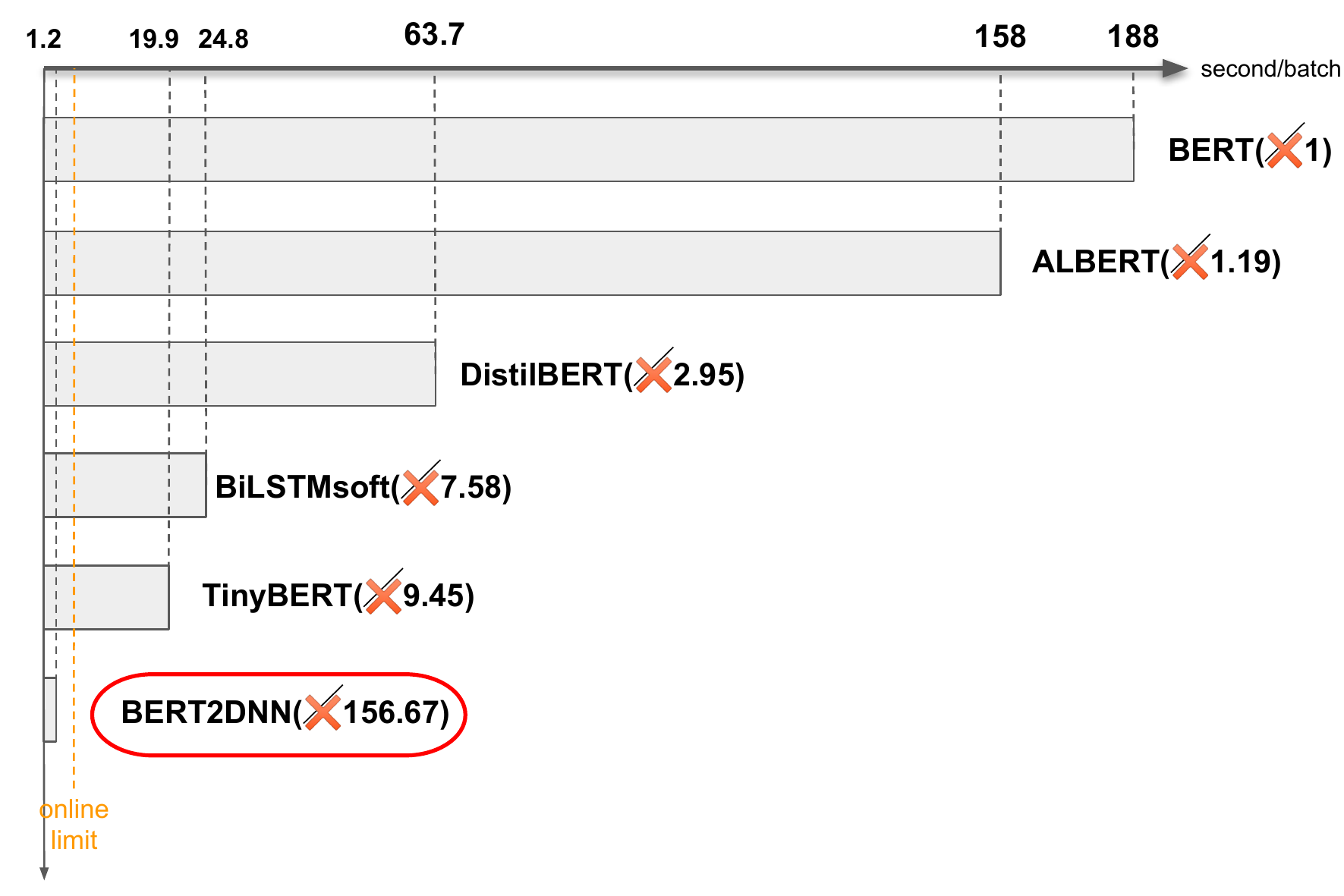}
\caption{Different distillation models' inference times on each batch. The red line denotes a time's limitation corresponding to the online response time's standard.} 
\label{time}
\end{figure}

To demonstrate the effectiveness of our approach, we conducted offline experiments and online A/B tests on in-house data. BERT2DNN achieves 7\% offline improvement on relevance metrics over previous deep neural transfer learning model\cite{jiang2019unified}. A/B tests on three types of sorting portals also show improvements on CVR and GMV related metrics. While our approach cannot exceed BERT in overall scores, the huge serving gain more than offsets the small relevance drop in industrial applications (Fig. \ref{time}). We further demonstrate the generality of our approach on a related public dataset, Stanford Sentiment Treebank (SST-2), by comparing with state of the art models and comprehensive studies on hyper-parameters. Additional case studies illustrate highly semantic relations of some learned embeddings and how well the student model imitates the excellent teacher model.

The major contributions of this paper are summarized as follows:
\begin{itemize}
\item We propose an effective learning framework with core techniques of transfer learning and knowledge distillation, in order to overcome the common problem of insufficient human labeled data in industrial applications.
\item Experiments and case studies demonstrate that simple feed forward student networks imitate the teacher models well, while bringing enormous boost in serving speed for online deployment.
\end{itemize}
\end{CJK}

\section{Related Work}
Relevance prediction is a well-studied area that ranges from the earlier work using GBDT (Gradient Boosting Decision Tree) based models to ones based on neural networks in the past decade. Its study in the e-commerce domain has been mainly carried out in the public by some of the better known e-commerce companies, most notably Alibaba, Amazon, Ebay, and JD.com. Here we outline a few notable works related to our own.

A typical commercial search engine consists of 2 or 3 phases, namely retrieval, coarse ranking, and optionally re-ranking. Since our method does not deal with retrieval, we will briefly survey historical developments in the latter categories.

\subsection{Traditional Ranking Methods}
Earliest ranking techniques include lexical bag-of-words matching and tf-idf scoring \cite{ramos2003using}. See \cite{li2014semantic} for a comprehensive survey. These however do not necessarily capture semantic similarity between query and document. Methods such as LSA and LDA can be viewed as precursors to the modern neural network based embedding approach, allowing non-exact matches. However they are typically unsupervised. 

For supervised relevance learning, we saw largely two camps prior to neural nets. In the linear model category, RankNet, RankSVM, and LambdaRank played prominent roles. In industrial settings, however GBDT remained popular for well over a decade and still is. It is best suited when features are mostly numeric. Subsequent developments include GBRank (pairwise) and LambdaMart (for list-wise objective). A comprehensive survey can be found in\cite{liu2009learning}. 

\subsection{Neural Network Models}

With the development of neural network, some deep models are gradually put into use in search engine. These approaches can mainly be divided into two categories: representation-based models and interaction-based models. As a classical representation-based model, Deep Structured Semantic Model (DSSM)\cite{huang2013learning} exhibits a structure of double towers, which is discriminatively trained by maximizing the conditional likelihood of the clicked documents given a query using the clickthrough data. The final relevance score between a query and an item is computed as cosine similarity of their corresponding semantic concept vectors. MatchPyramid\cite{pang2016text}, as a representative example of interaction-based models, constructs a matching matrix whose entries represent the similarities between words existing in query and document and uses a convolutional neural network to capture rich matching patterns in this matrix.

\subsection{BERT and Related Models}

BERT is a neural network model developed by Google. The release of BERT marks a singular breakthrough in the field of NLP because of its dramatic improvement over earlier states of art as well as its fine-tuning flexibility in a multitude of downstream tasks.

BERT's advent has also inspired many recent NLP architectures, training approaches and language models, such as Google's Transformer-XL\cite{dai2019transformer_xl}, XLNet\cite{yang2019xlnet}, Baidu's ERNIE\cite{zhang2019ernie}, Facebook's RoBERTa\cite{liu2019roberta}, etc.

A key factor in the success of BERT is the powerful role of Transformer structure, but Transformer itself has two defects, i.e., hard to capture long-term dependency and context fragmentation. Dai et al.\cite{dai2019transformer_xl} propose Transformer-XL structure which applies recurrence mechanism and relative positional encoding to address both of limitations. XLNet\cite{yang2019xlnet} integrates ideas from Transformer-XL into pre-training and achieves improved effects compared to BERT. Based on original BERT, ERNIE \cite{zhang2019ernie} designs a new pre-training objective by randomly masking some of the named entity alignments in the input text and asking the model to select appropriate entities from  knowledge graphs to complete the alignments. RoBERTa\cite{liu2019roberta} changes BERT in pre-training strategies by introducing dynamic masking and full-sentences so that it leads to better downstream task performance. 
ColBERT\cite{khattab2020colbert} applied BERT to encode query and item separately, before applying a late interaction function to estimate their relevance.

\subsection{Distillation Works}
Although pre-training models such as BERT and its variants have superior performance in NLP, it is difficult to deploy them in latency sensitive settings due to their enormous inference delays. Knowledge distillation\cite{Hinton2015Distilling} offers a potential solution to the deployment problem by means of model compression technique. Knowledge distillation usually chooses an expensive model as the teacher network and a small model as the student network. After the teacher network finishes training, it can be used to produce smoother simulated labels on a large number of unlabeled data. These model-annotated data are then used to train a student network that tries to mimic the generalization ability of the teacher network.

Some researchers have begun to explore BERT distillation\cite{jiao2019tinybert,sun2019patient,liu2019improving,sanh2019distilbert,tang2019distilling}. There exist two kinds of BERT distillation strategies: the first kind takes the same general architecture as BERT, i.e., Transformer, as the student network, the other chooses a different student network architecture (aptly named born-again network). TinyBERT\cite{jiao2019tinybert}, BERT-PKD\cite{sun2019patient}, MT-DNN\cite{liu2019improving}, DistilBERT\cite{sanh2019distilbert} all belong to the first category. They generally adopt three to six layers Transformers as the student network. In the second category, Tang et al.\cite{tang2019distilling} propose to distill knowledge from BERT into a single-layer BiLSTM, as well as its siamese counterpart for sentence-pair tasks. The above BERT distillation models indeed have high prediction accuracy, but according to our survey, these models still take 10\textasciitilde100 seconds to score 12800 examples. This makes it very difficult to deploy them on online server such as a modern search engine, which motivates us to design a more efficient BERT distillation solution.

\section{Model Description}
\subsection{Overview}

We use a simple 4-layer feed-forward network as the student model. For label, we use the BERT-Base prediction score on hundreds of millions to billions of query/item title pairs. The basic set of features for the student model is also query/title, and we choose a different tokenization scheme, namely Chinese \textbf{C}haracter and English \textbf{W}ord \textbf{U}nigrams and \textbf{B}igrams (\textbf{CWUB}). %In addition we convert all roman characters to lower case. 
Finally we retain about 3m highest frequency vocabulary total. This is about 100 times the vocab size of the BERT model. We find this to be an effective space/time complexity trade-off (Table~\ref{tab:eval-ai-result}).
% Nonetheless the total size of the vocab embedding (64-dimension) is still significantly smaller than even the BERT-Base model (768-dimension).

The model setup is similar to the one described in \cite{jiang2019unified}, with one major difference. In \cite{jiang2019unified}, pairwise training format \((query, item_a, item_b)\) was essential to learn from relative user preference signals: e.g., higher CTR or more click counts across two different queries do not necessarily mean more relevant results. In this work, we only need pointwise data since the score produced by BERT teacher model has absolute relevance meaning. This allows more compact aggregation of the training data, and avoids query distribution skew introduced by the pairwise expansion.

\begin{figure*}[htbp]
\centering
\includegraphics[width=0.85\textwidth]{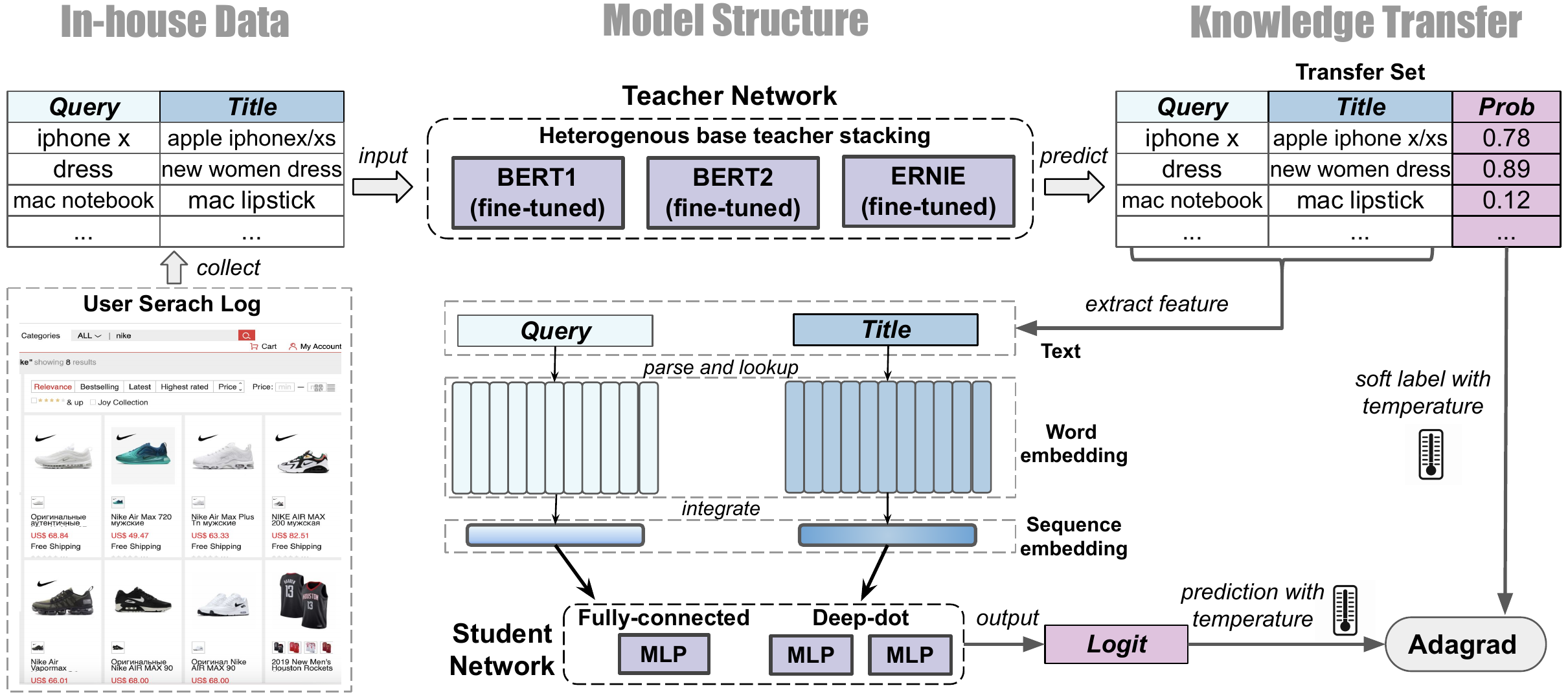} 
\caption{The knowledge transfer process of BERT2DNN. It mainly includes three parts: the use of in-house data, the designed distillation model structure and how to transfer knowledge from teacher model to student model.} 
\label{The knowledge transfer process of BERT2DNN}
\end{figure*}

We show the whole knowledge transfer process of the proposed BERT2DNN model in Fig. \ref{The knowledge transfer process of BERT2DNN} and illustrate it in detail in the following subsections. First we discuss how texts are converted to embeddings to be consumed by the models. We then describe our teacher and student model architectures in detail, comparing several candidates and techniques. Finally we describe how to transfer knowledge from teacher model to student model with temperature parameters.

\subsection{From Text to Sequence Embedding}

A typical NLP neural network model takes sentence embeddings as input. To prepare for the distillation step, we collect around 2 months of user search log, totaling ~170m query, item pairs. Following \cite{jiang2019unified}, we adopt the following tokenization procedure:

\textbf{CWUB}. Specifically, for texts with a mixture of Chinese and English, it uses the most basic semantic units, Chinese characters and lower cased alphabetic words, as unigrams. We do not normalize traditional and simplified Chinese since most of our texts are of the latter kind. Adjacent unigrams are further extracted to form bigrams, with the beginning and ending unigrams of each sentence forming additional bigrams with the special tokens ``\textasciicircum'' and ``\$''. Thus for instance, the query 
\begin{CJK*}{UTF8}{gbsn}
``mac电脑'' (mac computer) becomes
``\textasciicircum mac'', ``mac'', ``mac电'', ``电'', ``电脑'', ``脑'', ``脑\$''
\end{CJK*}.
% In addition, it converts all roman characters to lower case, as well as prepends ``\textasciicircum'' or appends ``\$'' to indicate beginning or end of sentence.
For the in-house dataset, we tokenize both query and item titles according to \textbf{CWUB}.
% According to \textbf{CWUB}, we first tokenize all the queries and item titles into Chinese characters and English unigrams, as well as adjacent bigrams, motivated by the fact that words and characters are the most natural semantic units in English and Chinese writing respectively. 

For learning efficiency as well as robustness, we filter unigrams and bigrams below a certain frequency level as observed in the training data, to reach a final vocab size of about 3 million. Lastly we initialize a 64-dimension vector for each vocab entry. The sentence embedding is constructed as
\begin{align}
\mathbf{e}_{s}=\sum_{i=1}^{n}\mathbf{e}_{w_{i}}/\sqrt{n}
\end{align}
where $S=\emph{w}_1, \emph{w}_2, ..., \emph{w}_\emph{n}$ are the unigrams and bigrams retained for the sentence, and $\textbf{e}_{\emph{w}_\emph{i}}$ is the embedding for word $\emph{w}_\emph{i}$.

\subsection{Model Structure}
The distillation process consists of two stages, the teacher phase and the student phase. The teacher model sets an upper bound on the model quality, whereas the student model is limited by online serving costs. Thus they have very different design principles discussed below.  

\subsubsection{Teacher model: from single-BERT to ensemble-model}
Inspired by the recent success of transfer learning in NLP tasks, especially Transformer network based models pioneered by Google-BERT, we chose the latter as our teacher model.

BERT and its subsequent refinements excelled at both single sentence and sentence pair classifications. In our e-commerce search setting, we focus mainly on sentence pair (query, item title) relevance.

Since our training data is dominated by (simplified) Chinese language material, the Chinese BERT-Base model becomes a natural choice for the teacher model. The only difference with BERT-Large is the number of Transformer layers: the latter uses 24 layers instead of 12. In Section~\ref{GLUE-experiment}, we will also test our ideas on an English corpus single sentence classification task.

For a sequence embedding of $n$ tokens and embedding dimension $d$, $X\in \mathbb{R}^{d\times n}$, the forward pass of one Transformer block including a self-attention layer Att(.) and a feed-forward layer FF(.) can be formulated as:
\begin{equation}
{\rm Att}(X)=X+\sum_{i=1}^{h}W_{O}^{i}W_{V}^{i}X\cdot \sigma [(W_{K}^{i}X)^{T}W_{Q}^{i}X]
\end{equation}
\begin{equation}
{\rm FF}(X)={\rm Att}(X)+W_{2}\cdot {\rm ReLU}(W_{1}\cdot {\rm Att}(X)+b_{1}1_{n}^{T})+b_{2}1_{n}^{T}
\end{equation}
where $W_{O}^{i}\in \mathbb{R}^{d\times m},W_{V}^{i},W_{K}^{i},W_{Q}^{i}\in \mathbb{R}^{m\times d},W_{2}\in \mathbb{R}^{d\times r},W_{1}\in \mathbb{R}^{r\times d}$, $b_{2}\in \mathbb{R}^{d}, b_{1}\in \mathbb{R}^{r}$, and FF($\emph{X}$) is the output of the Transformer block. The number of heads \emph{h} and the head size \emph{m} are two main parameters of the attention layer; and \emph{r} denotes the hidden layer size of the feed-forward layer.

BERT embeds a rich hierarchy of linguistic signals: surface information at the bottom, syntactic information in the middle, and semantic information at the top \cite{jawahar2019what}. Large model capacity and expressive power make BERT an ideal choice for the teacher model. To better align with the e-commerce relevance classification task, we fine-tune BERT on close to 400k human labeled query, item pairs. Concretely, we combine the query and item title tokens into a single sentence separated by the special token [SEP], prefixed by the special token [CLS], and padded at the end by [PAD]. We choose token length of 128 which covers more than 95\% of all our query, title pairs. The model proceeds by converting each token into 768 dimensional embedding, thus a total input size of 98304 per example (query, item pair). At the end, the model tries to minimize the cross entropy loss between the true label (relevant or not) and the output probability. We use mostly default hyper-parameters as suggested by the official model release: learning rate = 2e-5, epoch = 3, and batch size = 32.
On a separate hold-out set whose queries are not seen during training, the model reaches AUC of 0.8930 and Accuracy of 0.8606. These exceed the simple feed-forward model \cite{jiang2019unified} by 7-8\%, but serving latency also increases by more than 100 times. 

% TOOD(jyj): review stacking ensemble.
Single-BERT model often suffers from high variance on its test prediction scores. One way to mitigate that is through ensembling. By neutralizing individual model variance against one another, ensembling also postpones over-fitting on test data. We modify the stacking (ensemble learning) method described in \cite{zhou2012ensemble} to fit the distillation framework: 1. instead of base learners, first train multiple base \textbf{teachers} with the same training set ${\mathcal{D}}$, 2. collect prediction scores using the base teachers and generate new meta data ${\mathcal{D}}'$ (also called ``distilled data'' or ``transfer set''), 3. train a single meta learner based on the simulated label in ${\mathcal{D}}'$. Specifically, we design and experiment with two stacking distillation strategies:

\begin{algorithm}[H]  
\caption{Heterogeneous Base Teacher Stacking}  
\label{alg:Framwork}  
\begin{algorithmic}[1]  
\REQUIRE  
Data set $\mathcal{D}=\left \{(\mathbf{x}_{1},y_{1}),(\mathbf{x}_{2},y_{2}),\cdots,(\mathbf{x}_{Q},y_{Q})\right \}$; Heterogeneous base learning algorithms $\mathcal{L}_{1},\cdots ,\mathcal{L}_{K}$; Heterogeneous base teacher $h_{1},\cdots ,h_{K}$; Meta learning algorithm $\mathcal{L}'$; Meta learner ${h}'$. \\
\ENSURE 
$\mathit{H}(\mathbf{x})={h}'(h_{1}(\mathbf{x}),\cdots ,h_{K}(\mathbf{x}))$ \\
\FOR{\emph{k}=1,$\cdots$,\emph{K}}
\STATE Train base teacher $h_{k}=\mathcal{L}_{k}(\mathcal{D})$, then $h_{k}$ becomes base predictor
\ENDFOR \\
Define and initialize meta data ${\mathcal{D}}'=\varnothing$
\FOR{\emph{q}=1,$\cdots$,\emph{Q}}
\FOR{\emph{k}=1,$\cdots$,\emph{K}} 
\STATE Use base predictor to predict $z_{qk}=h_{k}(\mathbf{x}_{q})$
\ENDFOR\\
\STATE Define and initialize meta example's label $z_{q}=0$
\FOR{\emph{k}=1,$\cdots$,\emph{K}}
\STATE Accumulate prediction score $z_{q}=z_{q}+z_{qk}$
\ENDFOR\\
\STATE Calculate and generate meta example's label $z_{q}=z_{q}/K$ \\
\STATE Update meta data ${\mathcal{D}}'={\mathcal{D}}'\cup \left \{(x_{q},z_{q})\right \}$
\ENDFOR\\
Train meta learner ${h}'=\mathcal{L}'({\mathcal{D}}')$
\end{algorithmic}  
\end{algorithm}

\textbf{Homogeneous Base Teacher Stacking.}
Here we first train two BERT-Base models on the same human labeled data, which we name BERT1 and BERT2. While the training methodologies are identical, the two models can differ by virtue of random initialization and data shuffling. In the 2nd stage, prediction scores of BERT1 and BERT2 on user log data are combined by simple averaging to obtain ${\mathcal{D}}'$. Finally student model is trained directly on the averaged scores in ${\mathcal{D}}'$.

\textbf{Heterogeneous Base Teacher Stacking.}
Here we add prediction scores from ERNIE \cite{zhang2019ernie}, a close competitor of BERT. The main difference is that ERNIE used whole phrase masking in its pre-training step. This helps widening the gap between the two base learners. Now we simply average the prediction scores of all three BERT1, BERT2, and ERNIE together in stage 2, while the rest remains the same. We illustrate the heterogeneous stacking procedure in the Algorithm 1:

\subsubsection{Student model: fully-connected}
\begin{figure}
\centering
\subfigure[Fully-connected model]{
\includegraphics[width=0.35\textwidth]{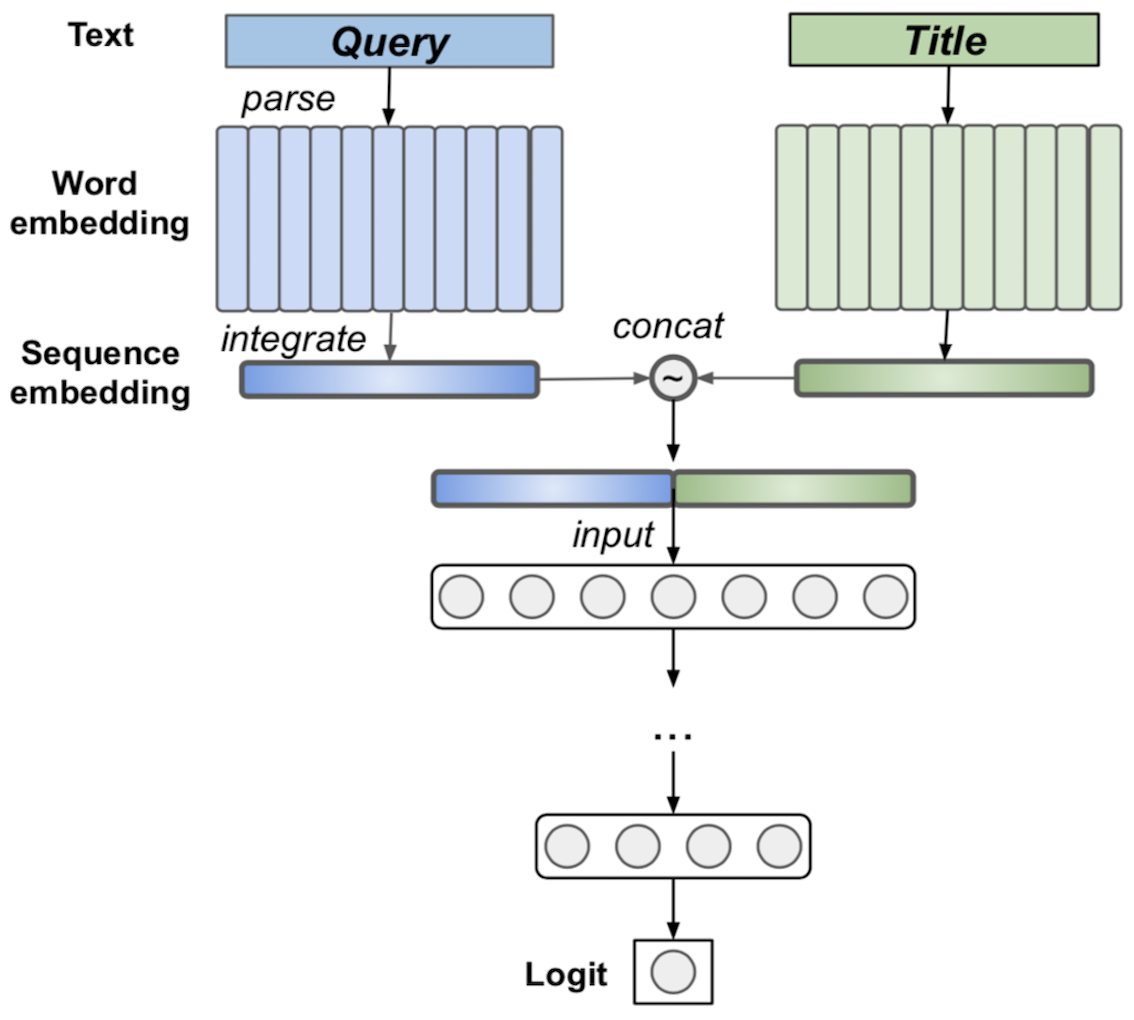}
\label{fully-connected}}
\hspace{0.01in}
\subfigure[Deep-dot model]{
\includegraphics[width=0.35\textwidth]{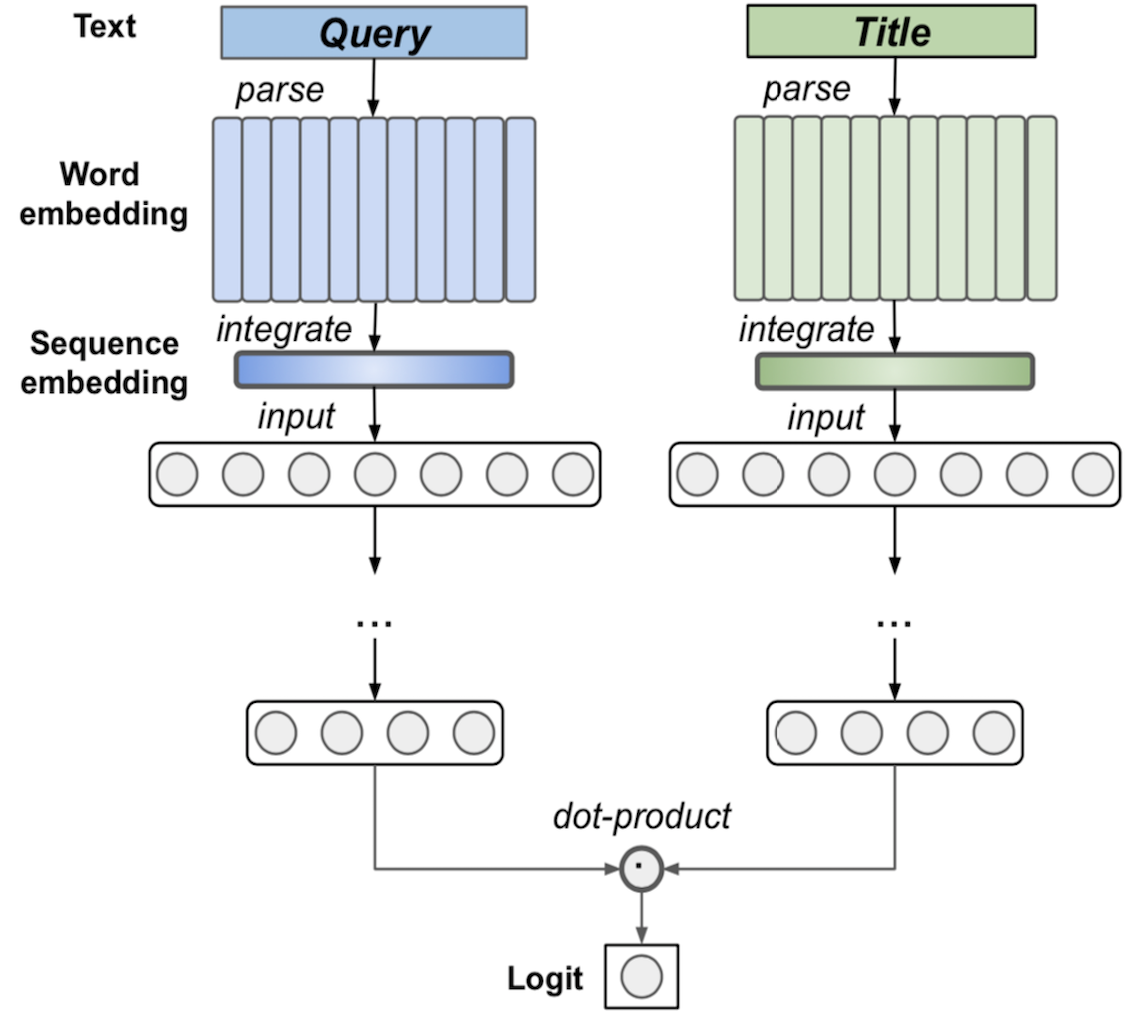}
\label{deep-dot}}
\hspace{0in}
\caption{Two types of feed-forward student models. Fully-connected model fully explores interactions of query and item, while deep-dot model produces better sentence representations.}
\label{student model}
\end{figure}

The student model (Figure \ref{student model}) takes aggregated token embeddings as sentence embedding input, and computes the score of a feed-forward Deep Neural Net (DNN). By the universal approximation theorem \cite{cybenko1989approximation}, DNN of just a single hidden layer can approximate any n-variate function under mild constraint. In addition, DNN has far less serving complexity and consumes much less computing resources than CNN, RNN, or Transformers. Therefore we focus on DNN as the student model of choice. 

Recall that the input layer is a 128 dimensional concatenation of query embedding and item embedding, each of 64 dimensions. As is typical in ranking applications, the fully-connected network consists of 4 hidden layers: [1024, 256, 128, 64], each activated by Rectified Linear Units (ReLUs). The last layer is mapped to a single logit value $\hat{y}$. Finally the loss for each example (\textbf{x}, \emph{y}) is computed via sigmoid cross entropy: 
\begin{equation}
L := -\left [ y\log(\frac{1}{1+e^{-\hat{y}}})+(1-y)\log(1-\frac{1}{1+e^{-\hat{y}}})\right ]
\end{equation}

\subsubsection{Student model: deep-dot}
Instead of mixing the query/item embeddings from the very beginning, it is also natural to consider a 2-tower model where the two output vectors are combined at the very end via a simple dot product (see e.g. \cite{huang2013learning}). Both query and item embeddings go through similar 4 hidden layers as in the fully-connected case, with different sets of parameters to accommodate the difference in embedding distribution. One advantage of the deep-dot model is the better representations of the embeddings for other tasks. Currently both versions of the student model have been deployed in production, under different product categories.
% For instance, the fully-connected model is applied to search results sorted by sale volume, whereas the deep-dot model has been applied to medical product search.

%\begin{figure}[ht]
%\centering
%\subfigure[Fully-connected model]{
%\includegraphics[width=0.21\textwidth]{3-a.png}}
%\hspace{0in}
%\subfigure[Deep-dot model]{
%\includegraphics[width=0.22\textwidth]{3-b.png}}
%\hspace{0in}
%\caption{Two different type student models}
%\end{figure}

\subsection{Distillation with Temperature}
In isolation, the BERT teacher model excels at prediction accuracy but suffers badly from slow inference and massive compute resource, while the student model architecture falls behind on prediction accuracy by a large margin. This motivates us to apply knowledge transfer from one to the other. From a practical stand point, the following conditions are essential for successful knowledge transfers:
\begin{itemize}
\item \textbf{Complete data resources}. The carrier data set must be sufficiently broad and abundant to cover all facets of the transfer task. Our data covers months of the full user search log, spanning top, torso, and tail queries and items of all major categories.
\item \textbf{Learning from soft labels}. Soft labels \cite{Hinton2015Distilling} containing model confidence information are preferred over hard binary labels. The former constitute a more natural learning target for the student model to imitate. In addition we apply different temperature parameters to construct several soft labels:
\end{itemize}
\begin{equation}
z_{i}^{T} = \frac{exp(z_{i}/{T})}{\sum_{j}exp(z_{j}/{T})}
\end{equation}
Here $z_i$ stands for the logit output of BERT. Higher temperature ${T}$ softens the label further, drawing positive and negative scores closer toward the middle. Experiment 5.1.2 shows our choice of temperature parameter that results in significant accuracy/AUC gains.

\section{Data Generation}
Our key innovation lies in the preparation of billion-scale data for the distillation process. The underlying hypothesis is that more data allows better approximation to a fine-tuned model since even though the latter is only trained on a small dataset of thousands to hundreds of thousands of examples, the pre-training stage has exposed the model to a much larger corpus. 
\subsection{Basic Distillation Data}
We prepare 4 sets of in-house data, all of which are divided into 90\% training and 10\% test set, based on queries, to avoid cross-leakage
\begin{itemize}
\item Human labeled \textbf{380k} (query, item) pairs with original relevance label in the range of 1 - 5. Editors are asked to judge the relevance of query and item and give a relevance score, with 5 as most relevant and 1 as most irrelevant. In our experiment, we binarize the original label, by regarding the relevance score as 1 for origin score of 4-5(positive label), while as 0 for 1-3(negative label).
\item 2 month of search log with query, item title pairs and user behaviors. The dataset is filtered by either ordered once or displayed 20 times without clicks, totaling \textbf{50m}. 
\item similar to the above but with more relaxed filtering criteria: clicked once or skipped at least 5 times, totaling \textbf{170m}. And skipped means shown by not click. 
\item 10 months of search log (query, item title) without any filtering, totally about \textbf{2.3b}.
\end{itemize}
The three sets of search log data provide additional data points to the data size comparison experiments in Fig. \ref{fig:distill_size}. 
\subsection{Additional Behavioral Data}
\label{behavior_data}
For online experiments, we make adjustment for the adverse effect of better relevance in user feedback, an example of which is provided by user typing the query ``iphone'' but ending up clicking on charging cable instead. In addition to the BERT scored search log data, we add query, item pairs that received user clicks as additional positive examples. This helps sway the model towards better recall for user clicks. 

\section{Experiments}
In this section, we present BERT2DNN experimental results on two tasks, which are search relevance on our in-house dataset and movie review sentiment classification on a public benchmark dataset: GLUE SST-2. The results on both dataset show the effectiveness of our proposed method and its capability to generalized to different tasks. For the in-house relevance quality task, we conduct evaluation on both offline and online metrics, showing the effectiveness on real world scenario. The experiment results on SST-2 dataset show BERT2DNN's capability of generalization to sentiment analysis task. And we also compare the model performance and inference time with other state-of-the-art models. 
\subsection{Relevance Quality on In-house Dataset}
We assess BERT2DNN framework with our in-house data and compare with some existing deep relevance models trained on CTR/CVR dataset.
%\footnote{We have prepared an experiment version of BERT2DNN code to release. It will take sometime to be available online due to company's policy.}. 
Here's a brief description of benchmark models.
%Within this section, we will refer to two kind of datasets, namely editor labeled dataset and CTR/CVR dataset. 
%\begin{itemize}
%\item CTR/CVR dataset is generated from JD.com log data, mainly containing query, item title, and cumulative amount of view, click, order. Both pointwise and pairwise data is used and the amount could be very large.
%\item Editor labeled dataset is a pointwise dataset, with each example contains query, item title, and a binary label representing if the query and item is relevant judged by editors. This dataset is splitted into train and test sets, and we use test set for all the results comparison. 
%\end{itemize}

\emph{DRFC} \cite{jiang2019unified}: Deep relevance model using fully-connected network. The model was trained in a 2 stages manner. First, a Siamese pairwise model was trained with shared parameters using click ratios as labels. The click model is then fine-tuned in a pointwise manner with 90\% of the 380k editor labeled data. 

\emph{DRDD}: Deep relevance model using deep-dot network. The data and training process are identical to DRFC. Instead of using a fully-connected deep neural network model, we calculate the inner product of the query embedding and item embedding as score. 
% For each baseline model, we illustrate the result of two versions: ``base'' and ``large''. The ``large'' model co-trains a large side network during fine-tuning, which is averaged with the main network at the end (see \cite{jiang2019unified} Figure 1(b)). 

% The `base' model is what we used, with a relatively fewer layers, while the `large' is a model with side network the same size as the fully-connected distilled model. We put it here simply for result comparison purpose. 

\emph{BERT/ERNIE Teacher}: BERT is fine-tuned from pre-trained Chinese base model\footnote{https://github.com/google-research/BERT}, using the 90\% training portion of the in-house editor labeled data. ERNIE is fine-tuned on the same data from pre-trained ERNIE 1.0 Chinese base model\footnote{https://github.com/PaddlePaddle/ERNIE}.

\emph{BERT2DNN}: This is a fully-connected student model distilled from BERT result. Unless other specified, default temperature at 1 is used to generate the student labels. 

\emph{BERT2Dot}: This is similar as BERT2DNN, but with deep-dot student model.

\emph{ERNIE2DNN}: The teacher model is ERNIE and student model is fully-connected model.

\emph{Homo./Heter. ensemble}: They are respectively homogeneous and heterogeneous stacking ensemble methods, as illustrated in Algorithm \ref{alg:Framwork}.

\begin{table}[ht]
\caption{Results on in-house data. Here we compare 13 models including 4 models trained on CTR/CVR data, BERT/ERNIE teacher model, 4 versions of student model and some ensemble methods. The best results of student models on 170m data are bolded.}
\label{in-house-offline}
\begin{tabular}{cccccc}
\toprule[0.8pt]
Model   & Acc    & Pr     & Rc      & F1      & AUC    \\ \midrule[0.8pt]
DRFC(large) & 0.833	& 0.851 &	0.950 &	0.898 &	0.821  \\ 
DRDD(large) & 0.823 &	0.860 &	0.920 &	0.889 &	0.816 \\
DRFC(base) & 0.815 &	0.874 &	0.888 &	0.881 &	0.822 \\
DRDD(base) & 0.811 &	0.868 &	0.889 &	0.879 &	0.814 \\
\hline
BERT Teacher & 0.861 & 0.884 & 0.943 & 0.911 & 0.893 \\
ERNIE Teacher & 0.860 & 0.897 & 0.924 & 0.911 & 0.892 \\
\hline 
BERT2DNN(50m) & 0.846 & 0.872 & 0.939 & 0.904 & 0.874 \\
\hline
BERT2Dot(170m) & 0.832 &0.845 &\textbf{0.959} &  0.898 & 0.837 \\
BERT2DNN(170m) & 0.847 &\textbf{0.879} &  0.930 &  0.904 & 0.870 \\
ERNIE2DNN(170m) &0.847 &0.875 &0.936 &0.904 &0.870 \\
Homo. ensemble(170m) &0.849 &0.871 &0.945 &0.906 &0.873 \\
Heter. ensemble(170m) &\textbf{0.853} &0.874 &0.945 &\textbf{0.908} &\textbf{0.883} \\
\hline 
BERT2DNN(2.3b) &0.855 &0.879 &0.942 &0.908 &0.883 \\
\bottomrule[0.8pt]
\end{tabular}
\end{table}

From the results shown in Table \ref{in-house-offline}, the BERT/ERNIE Teacher model achieves the overall best results. DRFC and DRDD represent previous best performers under feed-forward network architecture, but they are significantly worse than the BERT/ERNIE. We present the result of BERT2DOT (using deep-dot model as student model) and three versions of BERT2DNN trained with different distillation data sizes (50m, 170m, 2.3b). We track five relevance metrics: Accuracy, Precision, Recall, F1-score and AUC. The first four metrics are denoted as Acc, Pr, Rc, F1, respectively. As shown in Table \ref{in-house-offline}, as the amount of distillation examples increases, %in the transfer learning set, 
the model also performs better on test and at 2.3b examples gets close to BERT teacher model(AUC: -0.99\%, Acc: -0.67\%, F1: -0.24\%), but at a much lower inference cost.

\subsubsection{Ensemble model as teacher network}
\label{sec:ensemble}
To assess the impact of teacher ensembling on student model quality, we compare single teacher, as well as homogeneous and heterogeneous teacher ensembles. In both ensemble settings, we simply take the average of the various teacher prediction scores. In the homogeneous case, we fine-tuned two BERT-Base models on the human label data in an identical manner. The difference of the final teacher models comes mainly from the randomization of the training data. In the heterogeneous case, we added another teacher based on ERNIE \cite{zhang2019ernie}, which performs whole phrase masking during the pre-training stage. By itself the ERNIE teacher model performs similarly with BERT teacher. It is not surprising that averaging teacher predictions provide better student quality, since the teacher predictions themselves perform better on test when averaged. The addition of a heterogeneous teacher source, ERNIE, however, dramatically improves the student quality, on top of existing homogeneous ensemble. This suggests that the diversity of teacher labels plays an important role in distillation quality.

\begin{figure}[!ht]
\centering
\subfigure[In-house]{
\includegraphics[width=0.22\textwidth]{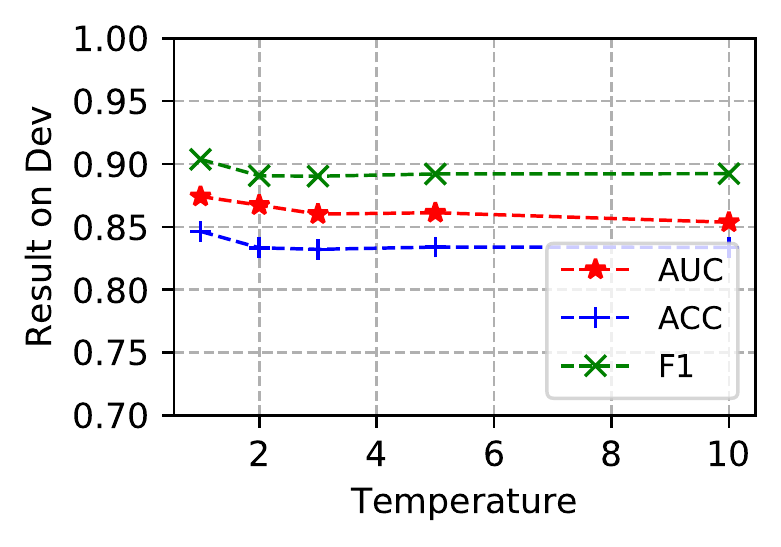}
\label{subfig:temp-inhouse}}
\hspace{0.05in}  
\subfigure[SST-2]{
\includegraphics[width=0.22\textwidth]{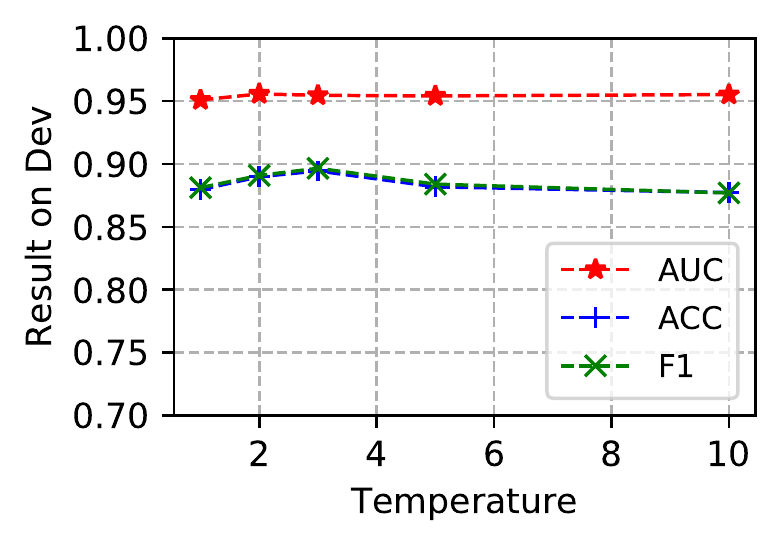}
\label{subfig:temp-sst}}
\caption{Impact of Distillation Temperatures}
\label{fig:temps}
\vspace{-20pt}
\end{figure}

\begin{figure}[!ht]
\centering
\includegraphics[width=0.4\textwidth]{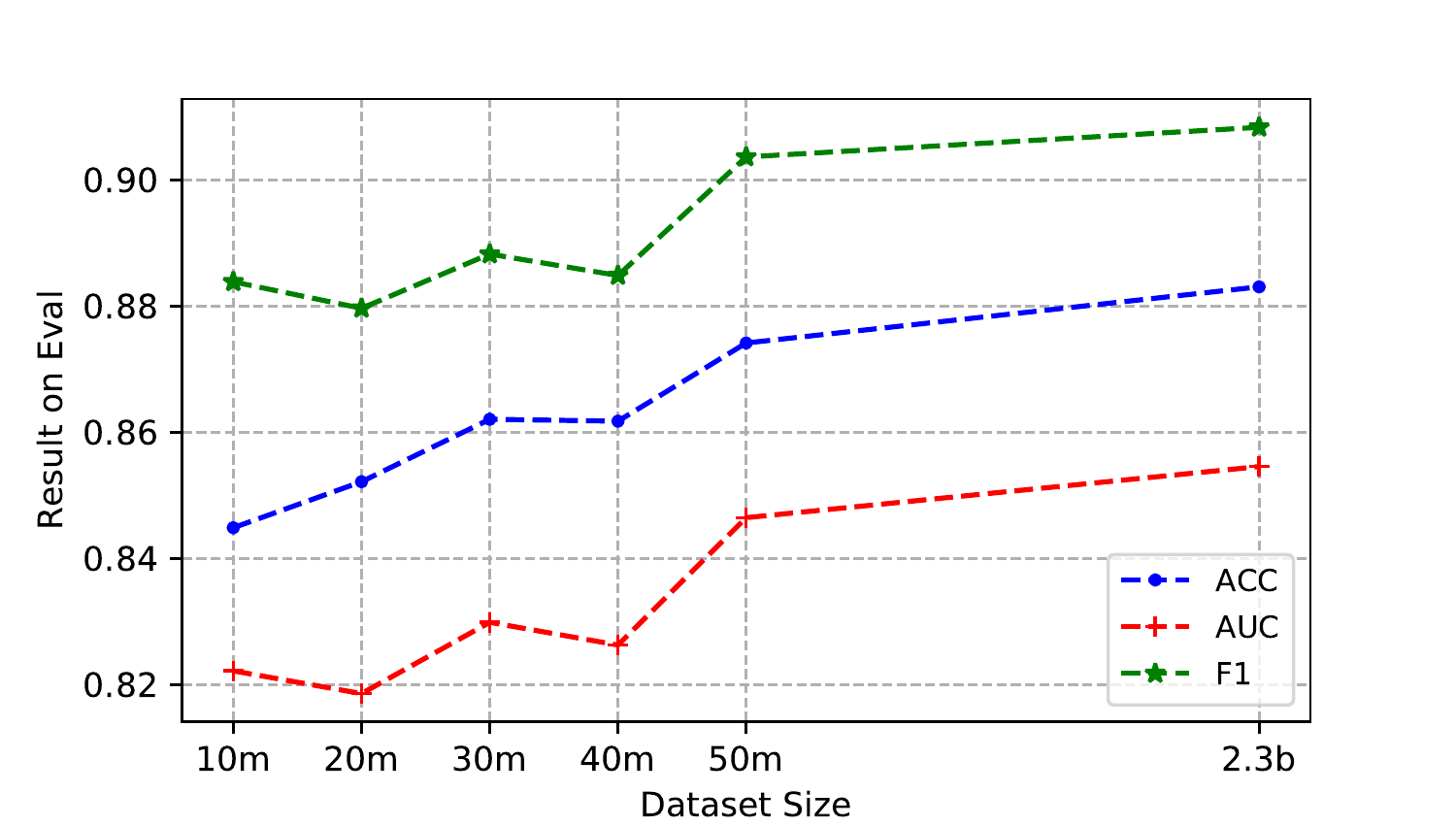} 
\caption{Impact of Distillation dataset size} 
\label{fig:distill_size}
\vspace{-10pt}

\end{figure}

\begin{table}[!ht]
\centering
\caption{Comparison of different vocabularies. The best results are bolded.}
\label{vocab_compare}
\begin{tabular}{ccccc}
\toprule[0.8pt]
Vocabulary   & Size    & AUC      & Acc      & F1  \\ \midrule[0.8pt]
BERT\_vocab  & 30522 & 0.723 & 0.791 & 0.879 \\
Our\_vocab  & 2476308 & \textbf{0.870} &
\textbf{0.847} & \textbf{0.904} \\
\bottomrule[0.8pt]
\end{tabular}
\end{table}

\subsubsection{The effect of different temperatures}
We investigate the effect of different temperatures for soft target `distillation', shown in Fig. \ref{fig:temps}. This study is conducted on both SST-2 task and our in-house experiment. For both tasks, we evaluate the performance using AUC, Accuracy and F1-score on \emph{T}(Temperature)=1,2,3,5,10 respectively. For the SST-2 task, temperatures ranging from 1-3 achieve good performances. Higher temperatures however confound the meta-labels too much and result in performance drops. For our in-house experiment, \emph{T}=1 beats other temperature settings. 

\subsubsection{Vocab size comparison}
The vocabulary set used in the student model is completely different from that in the teacher model. The teacher model uses WordPiece to generate about 30k tokens, about half of which are unigram Chinese characters and English letter n-grams, while the other half appends ``\#\#'' in front of the above tokens to indicate middle of sentence. The student model on the other hand takes a much rougher approach to get about 2.5m unigrams and bigrams based on 
\textbf{CWUB}. The overlap between the two vocab sets is about 10k, including almost all the Chinese unigrams in the BERT scheme.

To demonstrate the advantage of the larger vocab set, especially the use of bigrams, distillation experiment using the 30k teacher vocab is performed, with results shown in Table~\ref{vocab_compare}. We see that with the teacher vocab (BERT\_vocab), all major metrics drop significantly by an average of 5\%. This is not surprising as without the larger vocab set, the student model capacity would be significantly smaller than the teacher model.

\subsubsection{Distillation dataset size}
By subsampling from our distillation data of 2 months search log, we try to demonstrate the benefit of increasing the size of the transfer set. As show in Fig. \ref{fig:distill_size}, as the transfer data set increases from 10 m to 50 m, all the metrics(AUC, ACC, and F1-score) show an upward trend. And compared with these results, the 2.3b transfer set gained further improvement. The results also suggest larger transfer dataset with more coverage of the task context may be a one of the key of successful distillation model performance.

\begin{table}[!ht]
\centering
\caption{Online A/B testing experiments results.}
\label{in-house-online}
\begin{tabular}{ccccc}
\toprule[0.8pt]
Metrics &Sale sort &Price sort &Default sort \\
\midrule[0.8pt]
UV-value &+2.17\% &+8.82\% &+0.73\% \\
P-value &2.42e-2 &1.89e-2 &4.92e-2 \\
\midrule[0.8pt]
UCVR &+1.09\% &+0.18\% &+0.42\% \\
P-value &3.99e-3 &0.88 &3.54e-2 \\
\midrule[0.8pt]
UCTR &+0.09\% &+0.11\% &+0.12\% \\
P-value &0.57 &0.80 &5.20e-2 \\
\bottomrule[0.8pt]
\end{tabular}
\end{table}

\begin{table}
\centering
\caption{Results on SST-2 task dev dataset.}
\label{tbl:sst}
\begin{tabular}{cccccc}
\toprule[0.8pt]
Model &Acc &AUC &F1 &Pr	&Rc  \\
\midrule[0.8pt]
BERT Teacher & 0.931 & 0.977 & 0.927 & 0.938 & 0.926 \\ 
ERNIE Teacher & 0.939& 0.981&0.941&0.934&0.948 \\ 
\hline
DNN(label only) & 0.796 &0.862 &0.796&	0.804 &	0.793 \\
\hline
BERT2DNN & 0.880 & 0.951 & 0.881 & 0.874 & 0.892 \\
ERNIE2DNN & 0.905& 0.963 & 0.909 & 0.882 & 0.939 \\
Homo. ensemble & 0.891 &0.956&0.894&0.879&0.912 \\
Heter. ensemble &0.904& 0.958& 0.907 & 0.885 & 0.932 \\
\bottomrule[0.8pt]
\end{tabular}
\end{table}

\begin{table}
\centering
\caption{Inference performance comparison. The inference time of BERT2DNN (bolded) is minimum.}
\label{tab:eval-ai-result}
\begin{tabular}{ccccccc}
\toprule[0.8pt]
Model & Param  &L & Emb. & Time & Acc(dev/test)\\
\midrule[0.8pt]
BERT-Base\cite{devlin2019bert} & 109m & 12 & 768 & 188s & 92.9/93.5\\
MT\_DNNKD\cite{liu2019mt-dnn} & 109m & 12 & 768 & 188s & 94.3/95.6\\
ALBERT\cite{lan2019albert} & 12m & 12 & 128 & 158s  & 90.3/\hspace{0.09in}-\hspace{0.09in} \\
BERT\_PKD\cite{sun2019patient} & 52.2m & 4 & 768 & 63.7s & \hspace{0.09in}-\hspace{0.09in}/92.0 \\
DistilBERT\cite{sanh2019distilbert} & 52.2m & 4 & 768 & 63.7s & 90.7/\hspace{0.09in}-\hspace{0.09in} \\
BiLSTMsoft\cite{tang2019distilling}    & 10.1m    & 1 & 300 & 24.8s &\hspace{0.09in}-\hspace{0.09in}/90.7 \\
TinyBERT\cite{jiao2019tinybert} & 14.5m    & 3 & 312 & 19.9s  & \hspace{0.09in}-\hspace{0.09in}/92.6  \\
BERT2DNN & 158.8m & 4 & 64 & \textbf{1.2s} & 90.4/89.7
\\
\bottomrule[0.8pt]
\end{tabular}
\end{table}

\subsubsection{Online experiments}
Table~\ref{in-house-online} shows online A/B test metrics of BERT2DNN applied as a relevance filter to the baseline production search results, over a period of 2 weeks in JD.com. Our A/B test is conducted on three sort options, i.e. sale sort, price sort and default sort. As we stated, in such ranking options, relevance problem is much easier to reveal. Since relevance is often not well-aligned with user click/purchase preference (see section~\ref{behavior_data}), we added user behavioral data from each mode. This had little impact on the offline relevance metrics, but ensured good recall on historically clicked/purchased items. When the user issues a query in either search mode, production search results with model scores below a pre-chosen threshold are filtered out. In all search modes the filtering model produces generally positive behavioral trend, as seen by user view value (UV-value), unique user conversion rate (UCVR) and unique user click through rate (UCTR).

\subsection{Sentiment Analysis on GLUE SST-2 Subset}
\label{GLUE-experiment}

To demonstrate the generality of the distillation method, we apply our model on the task of Stanford Sentiment Treebank tasks(SST-2), which is one of the General Language Understanding  Evaluation(GLUE)\cite{wang2018glue} benchmark tasks for natural language understanding. SST-2 is a task of movie review sentiment classification. We pick SST-2 for evaluation since it's the task we could find additional unlabeled data for distilling knowledge to train student model. The additional data is Amazon Movie Review Dataset\cite{mcauley2013amateurs} from SNAP\footnote{https://snap.stanford.edu/data/web-Movies.html}, which consists of 7.9 m movie reviews, spanning a period of more than 10 years. Reviews include a plain text review and user ratings. We did a few basic data preprocess including remove HTML tags and sentence split, then received 64 m sentences.

\begin{figure}[!ht]
\centering
\subfigure[Fully-connected model]{
\includegraphics[width=0.35\textwidth]{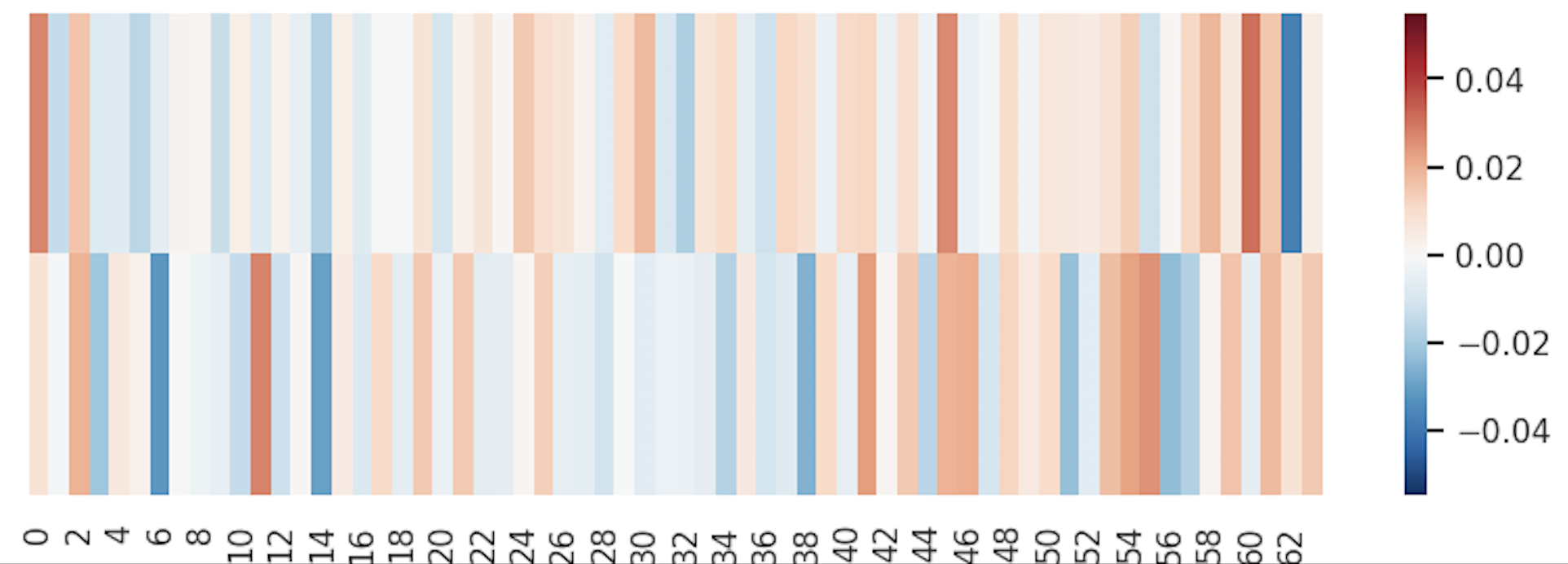}}
\hspace{0.2in}  %每张图片中间空闲
\subfigure[Deep-dot model]{
\includegraphics[width=0.35\textwidth]{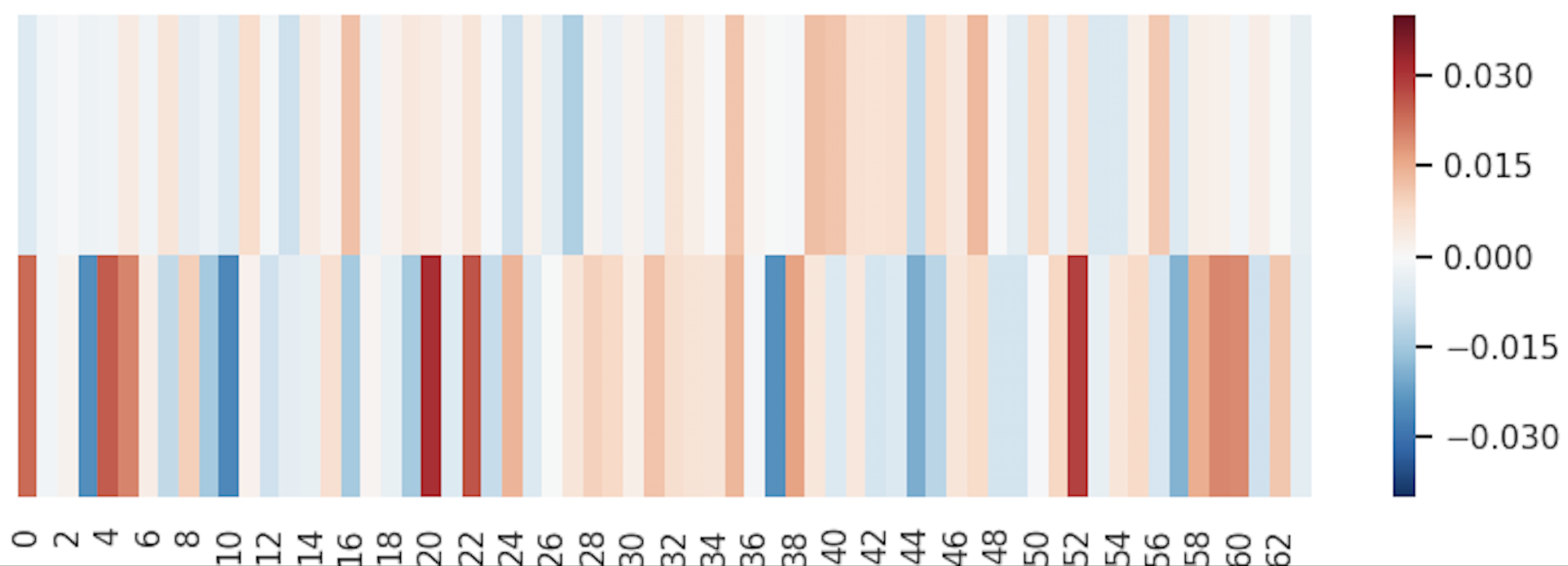}}
\hspace{0in}
\caption{Heatmap of sentence embeddings. The top/bottom of each subfigure represents ``red sweater'' and ``black sweater''/``mac computer'' and ``mac lipstick''.}
\label{heatmap}
\end{figure}

\begin{table}[t]
\centering
\begin{CJK*}{UTF8}{gbsn}
\caption{Distance of two sentence embeddings in fully-connected model (FC) and deep-dot model (DD).}
\label{case_embeddings}
\begin{tabular}{cccccc}
\toprule[0.8pt]
Case & Model &\multicolumn{3}{c}{Distance (Cos/Euc/Man)} \\ 
\midrule[0.8pt]
\multirow{2}*{\tabincell{c}{red sweater, black sweater}} &FC &0.208 &0.096 &0.606\\ 
&DD &0.126 &0.046 &0.293\\
\hline
\multirow{2}*{\tabincell{c}{mac computer, maclipstick}} &FC &0.392 &0.112 &0.734\\ 
&DD &0.604 &0.104 &0.671\\ 
\bottomrule[0.8pt]
\end{tabular}
\end{CJK*}
\end{table}

\begin{table}[ht]
\centering
\caption{Statistical metrics on teacher /student (T/S) model.}
\label{1000samples_stats}
\begin{tabular}{cccccc}
\toprule[0.8pt]
Mean (T/S) &Variance (T/S) &AUC &Acc &F1 &PCC\\
\midrule[0.8pt]
0.892/0.867 &0.053/0.057 &0.972  &0.957 &0.958 & 0.843\\
\bottomrule[0.8pt]
\end{tabular}
\end{table}

\begin{table*}[htbp]
\centering
\caption{Case of difference between teacher model and student model.}
\label{4cases}
\begin{CJK*}{UTF8}{gbsn}
\begin{tabular}{ccccc}
\toprule[0.8pt]
Query     & Title & Label & Teacher's score  & Student's score\\ 
\midrule[0.8pt]
men's coat &4M Jacket men's autumn/winter short pilot jacket black  &1   &0.996 &0.980\\
\hline 
grand piano &Japanese black vertical grand piano &1    &0.928 &0.916\\
\hline
slingshot & rubber band used in slingshot &0    &0.099 &0.024\\
\hline
HUAWEI x8 phone case & XINKE HONOR 7a/7c/7x phone case   &0    &0.244 &0.263\\
\bottomrule[0.8pt]
\end{tabular}
\end{CJK*}
\end{table*}

The overall results on dev set of SST-2 task are presented in Table \ref{tbl:sst}. BERT2DNN and ERNIE2DNN as student models, retain 94.6\%--96.3\% of the teacher model accuracy. DNN(label data) is trained directly on SST-2 train dataset only and has much lower accuracy than BERT2DNN.
% which demonstrates the necessity of using transfer dataset. 
Homo(geneous) ensemble model exceeds BERT2DNN accuracy by 1.25\%, showing the effectiveness of our teacher stacking strategy. Heter(ogeneous) ensemble model gains another 1.5\% accuracy on top of homogeneous ensemble, suggesting that BERT and ERNIE have complementary strengths as teacher models.
% achieves even higher accuracy retains 97\% accuracy of the teacher model, further deo the general effective of our model. 
Next We compare our method with several state of the art BERT related models on SST-2 task in Table \ref{tab:eval-ai-result}, together with parameter and inference latency analysis. We achieve very close accuracy on dev set as ALBERT and DistilBERT, while with simpler network structure and much lower serving cost. The performance in SST-2 task also shows the generality of the proposed BERT2DNN framework. 

\subsection{Prediction Speed Benchmarks}
As mentioned before, one great advantage of BERT2DNN is its ability to distill a computationally intensive model like BERT down to a light-weight feed-forward network. To validate the efficiency of the student model, we collect statistics on a range of BERT-inspired models, such as number of parameter/layers, embedding size, average inference time on 100 batches each of 128 examples, as well as Accuracy in SST-2 task, all in Table~\ref{tab:eval-ai-result}.

Besides our DNN student model and the 1-layer BiLSTM student model used in BiLSTMsoft\cite{tang2019distilling}, the remaining ones all use similar Transformer architecture as BERT. Despite the computational intensity, these latter models did not achieve breakthrough improvements compared to BERT.

Due to our large vocab size, the BERT2DNN model uses 50\% more parameters than BERT in serving (the training parameter size is almost identical because we used Adagrad instead of Adam). Indeed the bulk of the distilled DNN model comes from vocab embeddings (158.5m parameters), while the feed-forward part only takes up 0.3m parameters. Fortunately we manage to reduce inference time by several orders of magnitude, from the previous best record of 20s held by TinyBERT down to 1.2s. To facilitate reproducing the results on the SST-2 dataset augmented by SNAP dataset, we have released data processing and model training code\footnote{https://github.com/e-comerce-search/bert2dnn}. 
And our result offer another option for companies with no highly efficiency hardware to serve Transformer-based expensive models.

\section{Analysis and Case Studies}
The above experiments demonstrate the overall model quality and performance of BERT2DNN in terms of overall metrics. Next we analyze a few in-house examples to illustrate its qualitative strength. We hope to find out: what kind of sentence embeddings does BERT2DNN learn and how strong exactly is the student model.

\subsection{Sequence Embedding Analysis}
In this section, we look at sentence embeddings, formed by taking the sum of the constituent word embeddings. We chose a pair with similar semantics (``red'' and ``black''), and another pair of unrelated contexts (``mac computer'' and ``mac lipstick'').

We list Cosine/Euclidean/Manhattan distance for the two queries under fully-connected model and deep-dot model (Table~\ref{case_embeddings}). We use one minus cosine similarity of two sentence embedding vector calculated as cosine distance, for easier comparison with Euclidean/Manhattan distance. Note that both queries have more than 2 unigrams, making them unavailable as word embeddings directly. Compared to the fully-connected distillation student model, the deep-dot model displays a higher affinity between the two related sweater queries than between the two semantically unrelated ``mac'' queries. Indeed, Figure~\ref{heatmap} shows that the deep-dot model yields greater contrast in each dimension of the embedding vector for the two ``mac'' queries.

\begin{comment}
\begin{table}[t]
\begin{CJK*}{UTF8}{gbsn}
\caption{Distance of two sentence embeddings in fully-connected model (FC) and deep-dot model (DD).}
\label{case_embeddings}
\begin{tabular}{cccccc}
\toprule[0.8pt]
\multirow{2}*{Distance} & 红色毛衣/黑色毛衣 &  mac电脑/mac口红 \\ 
& (red/black sweater) & (mac computer/lipstick) \\ \midrule[0.8pt]
FC Cosine & 0.208 & 0.392\\
FC Euclidean & 0.096 & 0.112 \\
FC Manhattan & 0.606 & 0.734 \\
DD Cosine & 0.126 & 0.604 \\
DD Euclidean & 0.046 & 0.104 \\
DD Manhattan & 0.293 & 0.671 \\
\bottomrule[0.8pt]
\end{tabular}
\end{CJK*}
\end{table}
\end{comment}

\subsection{Score Differences between Student/Teacher Label}
A good distillation solution generally needs a high-accuracy teacher model and a strong-imitation student model. Here we show some real cases to detect the student model's imitation capability.

\begin{comment}
\begin{table*}[ht]
\centering
\caption{Case of difference between teacher model and student model.}
\label{4cases}
\begin{CJK*}{UTF8}{gbsn}
\begin{tabular}{ccccc}
\toprule[0.8pt]
Query     & Title & Label & Teacher's score  & Student's score\\ 
\midrule[0.8pt]
男装外套  &4M夹克男秋冬章仔短款飞行员夹克  黑色 165/88A   & \multirow{2}*{1}    & \multirow{2}*{0.996} & \multirow{2}*{0.980}\\
(men's coat) & (4M Jacket men's autumn/winter short pilot jacket black) &&& \\
\hline 
\multirow{3}*{\tabincell{c}{三角钢琴\\(grand piano)}}  &二手钢琴A+雅马哈钢琴YAMAHAUX300Wn   & \multirow{3}*{1}    & \multirow{3}*{0.928} & \multirow{3}*{0.916}\\
& 日本进口家用立式三角初学考级老师培训租售  黑色 &&& \\
& (2nd hand A+ YAMAHA piano ...  black) &&& \\
\hline
弹弓       &弹弓专用皮筋组 健身皮筋组拉力好  成品扁皮筋   & \multirow{2}*{0}    & \multirow{2}*{0.099} & \multirow{2}*{0.024}\\
(Slingshot) & (rubber band used in slingshot ...) \\
\hline
\multirow{3}*{\tabincell{c}{华为x8手机外壳\\(HUAWEI x8 phone case)}} &柯鑫 荣耀畅玩7a/7c/7x手机壳荣耀畅享8玻璃外壳后盖   & \multirow{3}*{0}    & \multirow{3}*{0.244} & \multirow{3}*{0.263}\\
& 炫光镭射8e全包防摔套抖音网红同款潮牌  小仙女 畅玩7X &&& \\
& (... XINKE HONOR 7a/7c/7x phone case ...) &&& \\
\bottomrule[0.8pt]
\end{tabular}
\end{CJK*}
\end{table*}
\end{comment}

We randomly select 1k cases for statistical analysis and relevance analysis. We calculate moment statistics of both the teacher and student prediction scores individually. The results are listed in Table \ref{1000samples_stats}. It is clear that the two models have very similar means and variances. We also calculate various classification metrics of student scores against teacher labels, including AUC, accuracy, F1-score and Pearson correlation coefficient (PCC). The fact that these are all very high suggests successful knowledge transfer from the teacher to the student model. In addition, we randomly select two positive cases and two negative cases from in-house data and report the ground-truth label and prediction score of teacher/student model in Table~\ref{4cases}. The prediction score of BERT2DNN is similar to that of BERT on all cases, suggesting that the distilled student model closely captures the teacher model's generalization skill.

\section{Conclusion}
We have demonstrated the capability of a simple low-complexity feed-forward network to achieve relevance quality similar to the state of the art BERT model, by distilling from the latter using a massive amount of data collected from an industry-scale user log. We showed the dramatic improvement in prediction speed, as well as the efficiency of the end to end training process. By augmenting a public dataset, we further validated the generality of our methods to other natural language scenarios. It is worth to explore ensemble learning and Automated Machine Learning (AutoML) on the student model, and we leave it in the future work.

\section*{Acknowledgment}
We are deeply indebted to Peng Zhao, Xuange Cui, Wei Si, Xianda Yang for strong online engineering support, and Wen-Yun Yang, Bin Li, and Zhaomeng Cheng for helpful modeling discussion. 
We thank the reviewers for many valuable suggestions.

\bibliographystyle{IEEEtran}
\bibliography{IEEEabrv,bert2dnn-main.bib}

\end{document}